\pdfoutput=1

\documentclass[11pt]{article}

\usepackage[]{emnlp2021}

\usepackage{times}
\usepackage{latexsym}
\usepackage{amsmath}
\usepackage{graphicx}
\usepackage{subcaption}
\usepackage{caption}
\usepackage{booktabs}
\usepackage{pifont}

\usepackage[T1]{fontenc}

\usepackage[utf8]{inputenc}

\usepackage{microtype}

\newcommand{\lin}{\textcolor{black}}

\newcommand{\revise}{\textcolor{black}}

\input{mySymbol.sty}

\title{Knowledge-Aware Graph-Enhanced GPT-2 for Dialogue State Tracking}

\author{Weizhe Lin \  \ \ Bo-Hsiang Tseng \  \ \  Bill Byrne\\
  Department of Engineering, University of Cambridge, United Kingdom \\
  \texttt{wl356@cam.ac.uk \ \  bht26@cam.ac.uk \ \ bill.byrne@eng.cam.ac.uk} \\
  }

\begin{document}
\maketitle
\begin{abstract}

Dialogue State Tracking is central to multi-domain task-oriented dialogue systems, responsible for extracting information from user utterances.
We present a novel hybrid architecture that augments GPT-2 with representations derived from Graph Attention Networks in such a way to allow causal, sequential prediction of slot values.
The model architecture captures inter-slot relationships and dependencies across domains that otherwise can be lost in sequential prediction.
We report improvements in state tracking performance in MultiWOZ 2.0 against a strong GPT-2 baseline and investigate a simplified sparse training scenario in which DST models are trained only on session-level annotations but evaluated at the turn level.
We further report detailed analyses to demonstrate the effectiveness of graph models in DST by showing that the proposed graph modules capture inter-slot dependencies and improve the predictions of values that are common to multiple domains.
\end{abstract}

\section{Introduction}
\label{sec:introduction}

This paper investigates two aspects of dialogue state tracking (DST) for multi-domain task-oriented dialogue~\cite{budzianowski2018multiwoz}.
We present a novel hybrid architecture that augments GPT-2~\cite{radford2019language} with dialogue act representations derived from Graph Attention Networks (GATs)~\cite{velivckovic2017graph} in such a way that allows causal, sequential prediction of slot values while explicitly modelling the relationships between slots and values across domains.
Our approach uses GATs to improve predictions of values that are shared across domain-slots and that might otherwise be treated independently.
As a related line of work, we investigate a form of sparsely supervised DST training and find that our hybrid architecture offers improved robustness with weak supervision.

DST can be improved by modelling the relationship between slots and values across domains.
This has been explored recently by \citeauthor{zhou2019multi} \citeyearpar{zhou2019multi} who suggest three types of relationships between domain-slots pairs that can be modelled explicitly:
(1) pairs that share the same candidate set, such as \texttt{<restaurant-bookday>} and \texttt{<hotel-bookday>};
(2) pairs whose candidate values are subsets, as could happen with \texttt{<restaurant-name>} and \texttt{<taxi-destina\\tion>} if the candidate set of the first belongs to that of the second; and (3) correlated values between domain-slot pairs, such as when the `star' level of a booked hotel correlates with the price range of a reserved restaurant.

Graph Neural Networks (GNNs) have been proposed to captures the interactions among slots and values and to improve DST performance \cite{zhou2019multi, 0002LWZT020, wu-etal-2020-gcdst}.  
These relationships can be represented as edges in graph-based models, where domains, slots, and values are nodes in the graphs.
However previous work has not explored quantitatively or in depth how graph models utilize the relationships they model.
\lin{\citet{0002LWZT020} and \citet{wu-etal-2020-gcdst} provide example cases where the predictions of correlated values were potentially enhanced by their model, while \citet{zhou2019multi} and \citet{zhu-etal-2020-efficient} present ablation studies showing marginal improvements brought by their graph modules.
\citet{zhu-etal-2020-efficient} and \citet{wu-etal-2020-gcdst} further show joint accuracies over different dialogue turns,
but there is more that can be said about how GATs can improve DST.}
One of the aims of this paper is to \lin{more deeply} analyze how graph models can lead to improved DST on top of an already good GPT-2 baseline system.
%

Graph models may also compensate for some potential drawbacks associated with using generative models for DST.
As a well-known generative model, GPT-2 offers powerful, left-to-right generation incorporating a causal attention mechanism.
We note that \citet{NEURIPS2020_e9462095} have demonstrated that GPT-2 can identify slot values as a prediction task, with variable length token sequences produced sequentially with interspersed special tokens indicating slot boundaries.
The ability to easily generate token sequences of arbitrary lengths is a valuable feature of the model,  although it may come at the expense of modelling power relative to models with non-causal attention mechanisms, such as BERT \cite{devlin-etal-2019-bert, shan-etal-2020-contextual}.
In particular, GPT-2's causality requires that the prediction of later slot values can depend explicitly on previously predicted slot values, but that the reverse is not possible. This can lead to decreased performance in predicting slot values that occur early on.
We find that augmenting GPT-2 prediction with representations derived from GATs allows some sharing of information between slots prior to prediction to improve this GPT-2 limitation.


Capturing the relationships of slot values across domains also offers the opportunity to make better use of limited training data, particularly in sparsely supervised and weakly supervised scenarios~\cite{liang2021attention}.
In a `Last Turn' annotation scenario, annotations are available only for the final turn of a task-oriented dialogue.
This is unlike the fully-annotated MultiWOZ setting, which offers turn-level annotations throughout the entire dialogue session.    
As an annotation option, generating summary annotations at the completion of a recorded session is an attractive alternative to creating a detailed, turn-by-turn annotation of the entire dialogue~ \cite{liang2021attention}.
If it is possible to use only these session-level annotations to train a DST system that still achieves acceptable tracking performance,  the chore of creating new annotated DST datasets could be made much easier.
The challenges in using this summary data are significant, however.   
Using only the final-turn annotations in MultiWOZ 2.0 reduces the training set to 14.3\% of its original size (in annotated turns).


We summarize the contributions of our work as follows:

(1) We propose a novel hybrid architecture that integrates GPT-2 with Graph Attention Networks (GATs) for dialogue state tracking.
The model is shown to be robust when training samples are significantly reduced under sparse supervision.

(2) We demonstrate that our architecture also mitigates a limitation of DSTs based on GPT-2 alone, associated with generating domain-slot values in a Left-to-Right manner.

(3) We investigate how knowledge-aware models capture relationships between domain-slots and show how using graphs can improve prediction of inter-dependent slot values.

While we do show DST accuracy improvements over a strong GPT-2 baseline, we emphasise that our aim is mainly to investigate and improve prediction of domain-slot values using relationships that otherwise are left unmodelled by the baseline. 





\section{Related Work}
\label{sec:related_work}

Statistical DST prioritises general and extensible systems based on machine-learning architectures~\cite{wu-etal-2019-transferable, zhang2019find, huang-etal-2020-meta, lee2020sumbt+}.
Systems must be able \lin{to} predict slot values from domain-specific lists \lin{such as list of hotel names} as well as from more open-ended categories such as days, prices, and times.
Recent trends are to combine several strategies to deal differently with the two types of values \citep{ zhang2019find, zhou2019multi, heck-etal-2020-trippy}.
For example, \citet{zhang2019find} combine a span predictor for non-enumerable slot values and a cosine similarity matching that exploits a BERT model to extract representations for enumerable slot values, with a dual-strategy model jointly handling both types of slot values;
\citet{zhou2019multi} use both a span predictor and a candidate classifier and combine their predictions with gating functions.
Our work is based on GPT-2 and we note that generative models such as GPT-2 are less widely used in DST tasks, possibly because they raise additional challenges for information aggregation subject to the causality, as discussed in Sec.~\ref{sec:introduction}. However these recent results show that these models can yield competitive DST accuracy \cite{NEURIPS2020_e9462095,yang2020ubar} .

Previous work has addressed sharing  information between slots either by explicitly copying values~\cite{ouyang-etal-2020-dialogue,heck-etal-2020-trippy} or by sharing embeddings \cite{hu-etal-2020-sas, zhou2019multi, 0002LWZT020}.    Beyond copying and sharing, as we note in Sec.~\ref{sec:introduction}
\citet{zhou2019multi} developed a graph attention network, and \citealp{0002LWZT020} also developed a schema-guided multi-domain approach embedding slot relations in edges of graph neural networks.
\lin{\citet{zhu-etal-2020-efficient} enhanced a strong base model SOM-DST \cite{kim-etal-2020-efficient} with a schema graph to exploit relations among domain-slots.
GCDST~\cite{wu-etal-2020-gcdst} uses a state graph to transfer domain-slot features and hard-copy states directly from historical states.
}

\section{Graph Neural Networks}
In this section we review Graph Attention Networks (GATs) \cite{velivckovic2017graph, li2021message} as will be used in this paper.

A weighted \lin{undirected} graph at each dialogue turn $t$ is defined as
$\ccalG = (\ccalV, \ccalE)$ with a node set $\ccalV$ consisting of $N$ nodes $\{v_{i}\}$,
and an edge set $\ccalE$ containing all edges between nodes.
We define an $N\times N$ binary symmetric adjacency matrix $\bbS$, where $[\bbS]_{ij}  = 0$ if $(v_{i},v_{j}) \notin \ccalE$ and $1$ otherwise.
Associated with each node $v_i$ are feature vectors $\bbx^i \in \mathbf R^F$.
These are gathered into matrices $\bbX$ of dimension  $N\times F$, \lin{where $F$ is the input feature size.}

\begin{figure}
    \centering
    \includegraphics[width=0.7\linewidth]{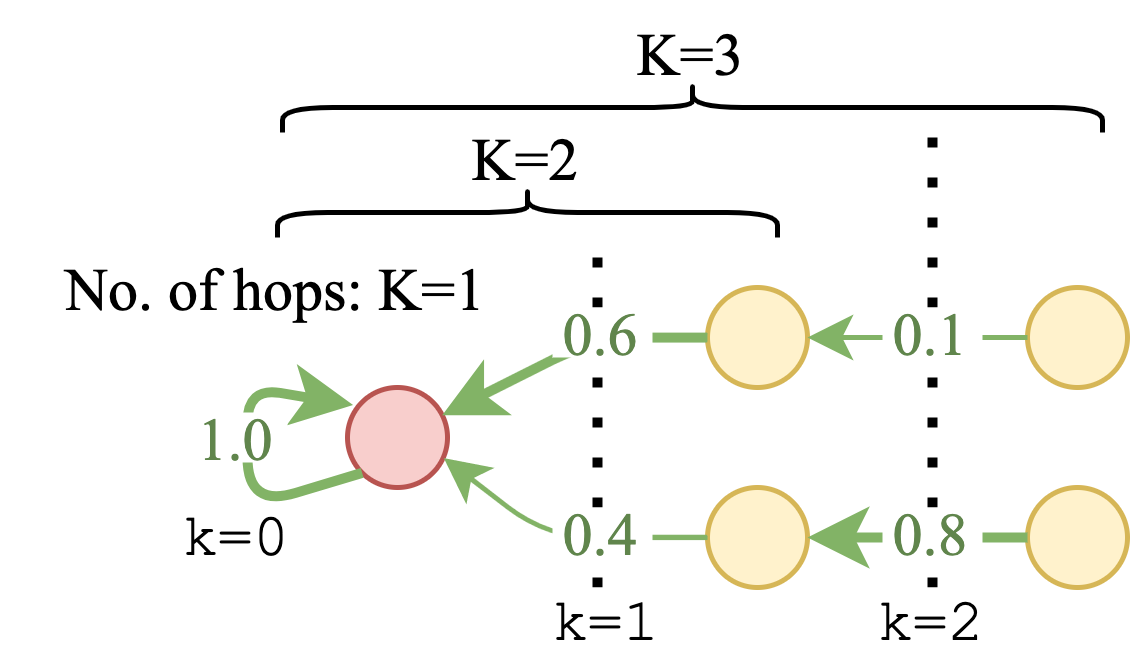}
    \caption{Illustration of GATs. $k=0$ is self-connection, and $k\geq 1$ passes the features of other nodes to the node being evaluated.
    The values on the links are attention values, which weight the passing features.}
    \vspace{-0.5cm}
    \label{fig:gat_sketch}
\end{figure}

Note that $\bbS\bbX$ is mathematically equivalent to passing the features of each graph node to its neighbours.
In this way $\bbS^{k} \bbX = \bbS(\bbS^{k-1}\bbX)$ is equivalent to $k$ rounds of feature exchanges with neighbours.
As illustrated in Fig.~\ref{fig:gat_sketch}, $k=0$ is self-connection, while $k>0$ aggregates features from $k$ nodes away.

A GAT layer transforms an input $\bbX \in \mathbf{R}^{N\times F}$ to an output $\mathcal G(\bbX) \in \mathbf{R}^{N\times G}$ as follows.
Each $K$-hop GAT layer consists of $P$ attention heads $\ccalA^{(p)}$ which incorporate $k=0,...,K-1$ rounds of feature aggregation (as shown in Fig.~\ref{fig:gat_sketch}) across the graph as
\vspace{-0.5cm}
\begin{equation}
\label{eqn:graphAttentionConv}
\begin{split}
    \ccalA^{(p)}(\bbX; \bbS) & = \sum_{k=0}^{K-1} (\bbE \odot \bbS)^{k} \bbX \bbA^{(p)}_{k}  \\
    \mathcal G(\bbX) &  =  \frac{1}{P}\sum_{p=1}^{P}    \sigma \big[ \ccalA^{(p)}(\bbX;\bbS) \big], 
\end{split}
\vspace{-0.5cm}
\end{equation}
where the $\{\bbA_{k}^{(p)}\}_{k=0}^{K-1}$ are $\mathbf{R}^{F \times G}$ linear feature transforms
and $\sigma(.)$ is a non-linear activation function.
The values of the $N\times N$ attention matrix $\mathbf E$ are computed
over $\bbX$ as 
\begin{equation} 
\begin{split}
    [\mathbf{E}]_{ij} & = \frac{\exp{(LeakyReLU(e_{ij})})}{\sum_{k\in \ccalN_{i}}\exp{(LeakyReLU(e_{ik})})}  \\
    e_{i j} & = (\bbx^{i})^{\top} \boldsymbol{Q}^{(p)}\bbx^{j} ,
\end{split}
\label{eqn:GAT_attention}
\end{equation}
where $\ccalN_i$ are the neighbouring nodes of node $v_i$, and $\boldsymbol{Q}^{(p)}$ are trainable $F\times F$ matrices used in computing attention.
In this way a GAT layer aggregates features selectively by assigning dynamic weights to graph edges based on the input node features.

GATs are formed as a cascade of $L$ GAT layers $\mathcal G_\ell $, each with its own multi-headed graph attention mechanisms ${\ccalA}^p_\ell$.
\lin{At time $t$,}  the GAT transforms a set of input features  $\bbX_t^{(0)}$ 
to a set of output features $\bbX^{(L)}_t$ as 
\begin{equation} 
    \bbX^{(\ell)}_t =  \mathcal G_\ell (\bbX_t^{(\ell-1)}) \text{ for } \ell = 1,\ldots, L
\end{equation}
Note that in this paper, we set output dimension $G=F$ for all GAT layers such that the GAT output features have the same dimensions as the input.


\section{Dialogue State Tracking with GPT-2 and Graph Neural Networks}

\begin{figure*}
    \centering
    \includegraphics[width=\textwidth]{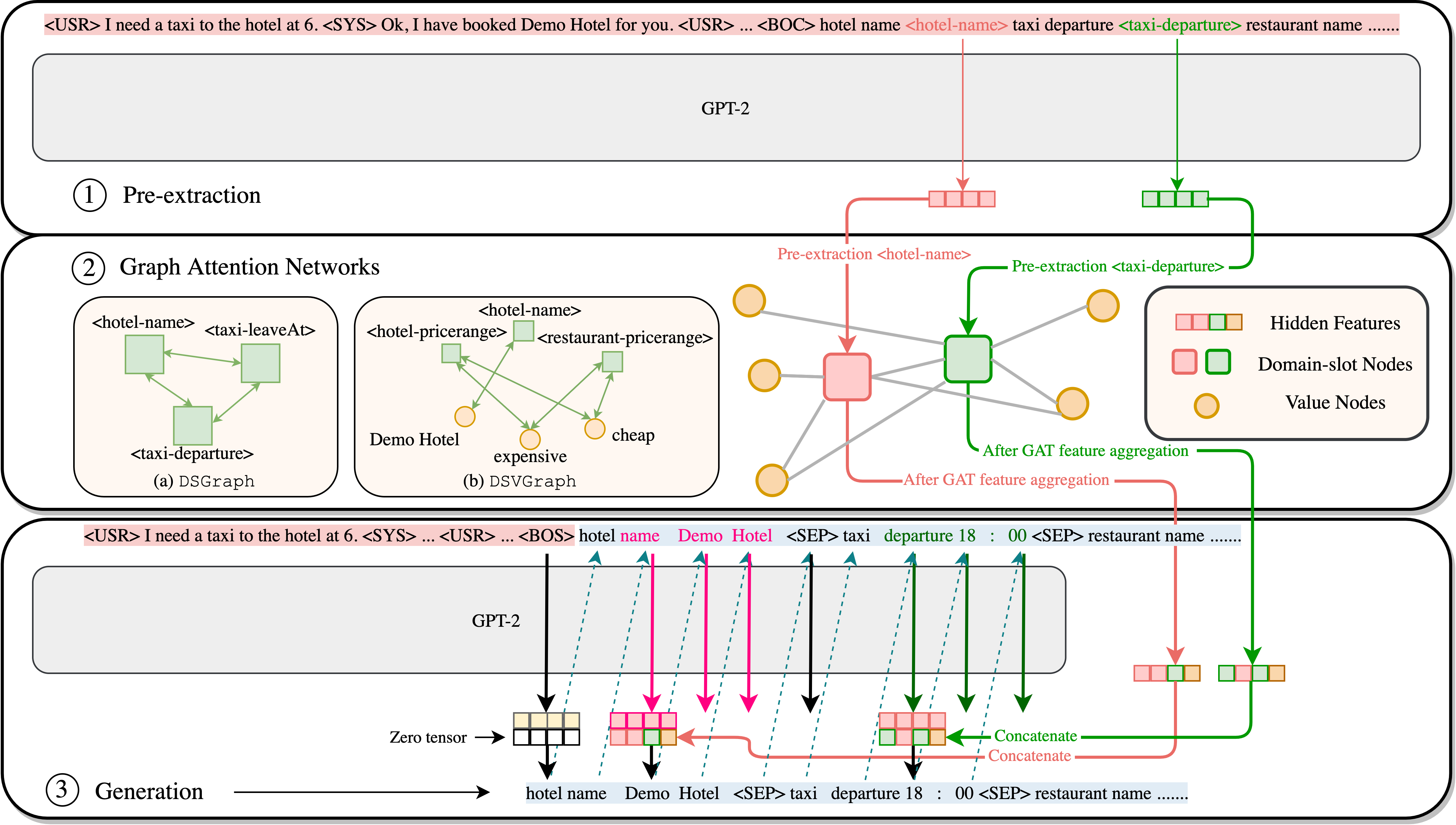}
    \caption{The workflow of the proposed model: 
    \ding{172} The model extracts domain-slot \lin{embeddings} from dialogue history, without knowing the ground truth; \ding{173} domain-slot \lin{embeddings are} passed into Graph Attention Networks for feature aggregation and information exchanges; \ding{173}(a)-\ding{173}(b) two types of graph connectivity used in our experiments; \ding{174} the updated domain-slot \lin{features are fed into the causal generation process of corresponding slots.} 
    Tokens shaded with red are model inputs, while tokens shaded with blue are generation outputs.
    For better visualization, only two domain-slot pairs are presented (\texttt{<hotel-name>} and \texttt{<taxi-departure>}).}
    \label{fig:workflow}
    \vspace{-0.4cm}
\end{figure*}


We take a three-step approach to incorporating GNNs into GPT-2 for dialogue state tracking (see Fig~\ref{fig:workflow}).    
At each turn we first present GPT-2 with the  dialogue history to generate features for all possible domain-slots and values in the ontology.
These features are then \lin{fed into} a GAT which captures relationships amongst domain-slots and values.   The features produced at the output layer of the GAT are then incorporated into a second application of GPT-2 which performs the actual prediction of the dialogue state values.

\subsection{Domain-Slot and Value Embeddings}
\label{sec:pre_extraction}
\lin{The first step is to extract features of both domain-slots and values in the ontology.}
The dialogue history $H_t$ at turn $t$ is a concatenation of user utterances and system responses, separated with special tokens: 
$H_t$  =  `$u_t$ \texttt{<SYS>} $s_{t-1}$ \texttt{<USR>} $u_{t-1}$ ... \texttt{<SYS>} $s_{1}$ \texttt{<USR>} $u_{1}$'.
From the ontology, we construct a string for all domain-slots as follows:
$F =$ {\em `hotel name \texttt{<hotel-name>} taxi departure \texttt{<taxi-departure>}...}'~.
The string $F$ contains all domain-slots in the ontology \lin{and does not change with samples}.
The domain-slots appear in a fixed order and each is preceded by a brief text description to  provide context to  GPT-2 in producing features.

To produce domain-slot features at dialogue turn $t$,  the string {\em `$H_t$ \texttt{<BOC>}  $F$'} is presented to GPT-2.
\lin{Since} the domain-slots are fixed and appear in a prescribed order in $F$,  there is a straightforward link between the positions of domain-slots in the input and their embeddings in the GPT-2 output layer.
For example, the feature for \texttt{<taxi-depature>} can be found in the same position of the output embedding sequence as that domain-slot appears in the input, \lin{as shown by arrows in Fig.\ref{fig:workflow} \ding{172}: Pre-extraction}.

To produce embeddings for all possible values in the ontology at turn $t$
the embedding layer of the GPT-2 is used.
\lin{Therefore, this representation is fixed from turn to turn until the embedding layer is updated in back propagation.}
Some values may consist of multiple tokens, e.g. {\em `Demo Hotel'} for the domain-slot \texttt{<hotel-name>}.
A single vector for each multi-token value is found by averaging the features of each token.

At dialogue turn $t$, the domain-slot features and the value features  are gathered into matrices $X^s_t \in \mathbf{R}^{N_s \times h}$ and 
$X^v_t \in \mathbf{R}^{N_v \times h}$,  where there are $N_v$ values and $N_s$ domain-slots, and $h$ is the size of the hidden layer of the GPT-2 Transformer.

\subsection{Inter-slot Information Exchange}
\label{sec:gat_layers}

We will use two types of GATs: \texttt{DSGraph} and \texttt{DSVGraph}.
In \texttt{DSGraph}, there are $N_{s}$ nodes, each representing a domain-slot pair. All nodes are connected to each other to allow nodes to exchange features as shown in Fig.\ref{fig:workflow} \ding{173} (a).
In \texttt{DSVGraph}, there are $N_{s}$ domain-slot nodes and $N_v$ value nodes, each of the latter representing a possible value.   If a value is in the candidate set of a domain-slot pair, then the corresponding value node and domain-slot node are connected, as shown in Fig.\ref{fig:workflow} \lin{\ding{173}} (b).    The domain-slot nodes are not otherwise  connected.

With features for domain-slots and values extracted in Sec.~\ref{sec:pre_extraction},  we use GATs to transform the features to capture the relationships between domain-slots and values.  The inputs to the GATs are    
$$
\bbX_t^{(0)}=\left\{
             \begin{array}{lr}
             X_{t}^{s} \in \mathbb{R}^{N_{s} \times h} &  in\ \texttt{DSGraph},\\
             X_{t}^{s} || X_{t}^{v} \in \mathbb{R}^{(N_{s}+N_v) \times h} & in\ \texttt{DSVGraph}
             \end{array}
\right.
.
$$
We use only the resulting domain-slot embeddings after graph \lin{operations}, and thus we extract the first $N_{s}$ items of the output tensor $\bbX^{(L)}_t$ and gather them into a matrix $\mathbf G_t \in \mathbb{R}^{N_{s}\times h}$.


\vspace{-0.2cm}
\subsection{Dialog State Prediction}
\label{sec:prediction}

\lin{Finally,} we present the string `$H_t$ \texttt{<BOS>}' to the GPT-2 model to predict the dialogue state.
The model is required to generate output $Y_t$, a sequence of tokens of serialized domain-slot pairs and corresponding values: $Y_t=$ {\em `hotel name Demo Hotel} \texttt{<SEP>} {\em taxi departure 18 : 00} \texttt{<SEP>} {\em ...} \texttt{<EOS>}'. 
The model is trained to generate the name of each domain-slot,  its predicted value, and finally a separation token \texttt{<SEP>} before proceeding to the prediction of the next domain-slot.
Note that the value `none' is generated for empty/not mentioned domain-slot values and thus all slots will be generated regardless of whether they have values.
After producing values for all domain-slots, the model generates an \texttt{<EOS>} to end the generation process.
In practice, we find that the model never omits any of the $N_{s}$ domain-slot pairs during generation,  further confirming GPT-2's ability to produce structured output.
An example of input/output is at the bottom of Fig.\ref{fig:workflow} \ding{174}: Generation.

\textbf{Decoding:} In generation the model incorporates the GAT features $\bbG_{t}[i] \in \mathbb{R}^{h}$
as shown by the pink arrows in Fig.\ref{fig:workflow} \ding{174}: Generation.  When predicting the value of the $i^{th}$ domain-slot (in this example the domain-slot is \texttt{hotel-name}), the GPT-2 features used for token decoding are concatenated with the domain-slot features $\bbG_{t}[i]$ from the output of the GATs.
The prediction of the value for each domain-slot will incorporate the domain-slot features produced by the GATs.
When predicting tokens that are not related to value predictions (black arrows in the figure), an all-zero tensor is concatenated to keep consistency.
The input dimension of the linear layer in decoding is extended to accommodate the GAT features in concatenation.

{\bf Fine-tuning:} Fine-tuning of GPT-2 for MultiWOZ is done in the usual way.   Each turn $t$ in the MultiWOZ training set is transformed into a sequence `$H_t$ \texttt{<BOS>} $Y_t$' where $Y_t$ contains the sequence of domain-slots and values for dialogueturn $t$, as extracted from the annotated training set.    Training proceeds by optimising $P(Y_t | H_t; \theta)$ over the training set.

\section{Experiments}
\begin{figure*}[!ht]
    \centering
    \includegraphics[width=\linewidth]{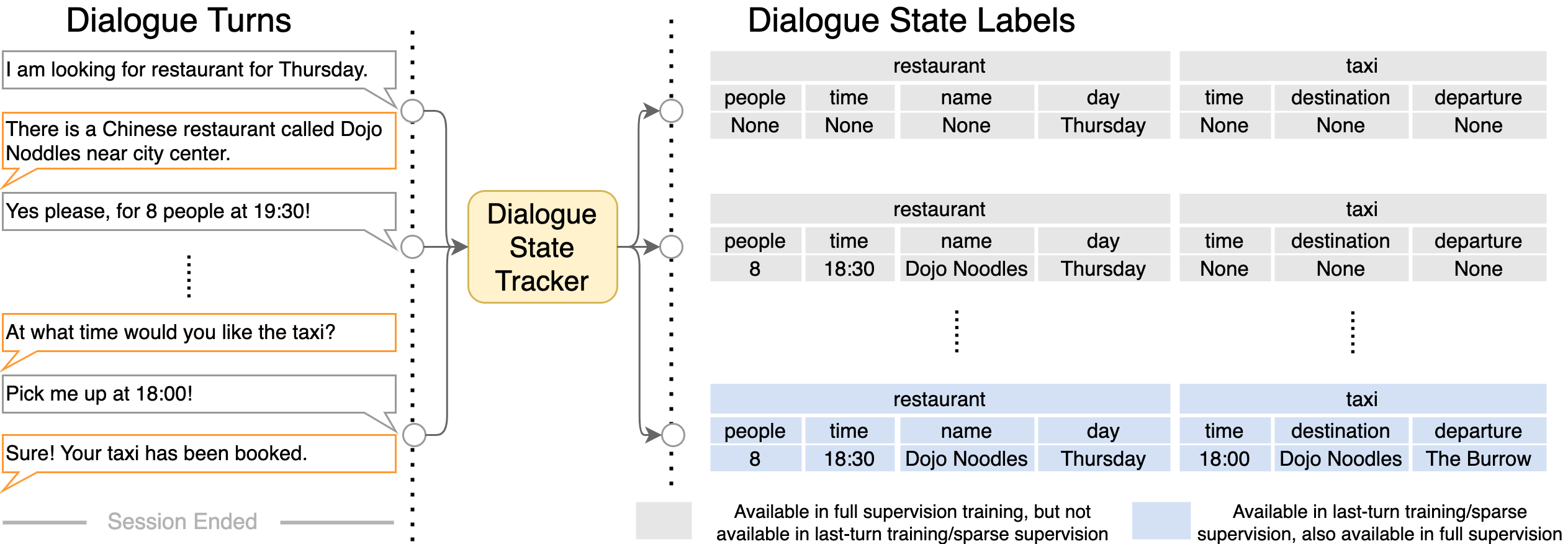}
    \caption{
    \revise{An example of the sparely-supervised training scenario where only the annotations at the last turn (highlighted in blue) are available.}
    }
    \label{fig:DST_example}
\end{figure*}
We report dialogue state tracking performance on \textit{MultiWOZ 2.0}~\cite{budzianowski2018multiwoz} with its multi-domain goal-oriented dialogue conversations and annotations. 
For direct comparison to the previous literature, we use the same preprocessing as  \citeauthor{wu-etal-2019-transferable} \citeyearpar{wu-etal-2019-transferable} and \citeauthor{zhou2019multi} \citeyearpar{zhou2019multi}.
Two metrics are used for evaluating the model performance:

\textit{Slot Accuracy} measures the ratio of successful slot value predictions among all the slots of each dialogue turn in ground-truth.

\textit{Joint Goal Accuracy} compares the predicted \lin{belief state} to the ground truth at every dialogue turn,  and the output is considered correct only if all the predicted slot values exactly match the ground truth values.

\subsection{Baseline Performance}

\begin{table*}[h!]
\centering
     \begin{subfigure}[b]{0.38\textwidth}
\centering
\begin{tabular}{lrr}
\toprule
                 & \multicolumn{1}{l}{\small{Joint (\%)}} & \multicolumn{1}{l}{\small{Slot (\%)}}  \\
                 \midrule
TRADE            & 48.62                              & 96.92                              \\
DSTQA*            & 52.24                              & 97.28                              \\
SimpleTOD*        & 51.37                              & 96.48                                   \\
SOM-DST+SG & 52.53 & N/A \\
GCDST & 50.68 & N/A \\
SST & 51.17 & N/A \\
SUMBT+LaRL       & 51.52                              & \textbf{97.89}                              \\
DST-Picklist     & 54.39                              & N/A                                \\
L0P0K0-NoGraph   & 53.00                                 & 97.34                              \\
L4P4K2-DSGraph & \textbf{54.86}                              & 97.47                              \\
L4P4K2-DSVGraph & 54.62                                & 97.42                               \\
\bottomrule
\end{tabular}
         \caption{Training with all samples.} \label{tab:compare_baselines}
     \end{subfigure}
     \hspace{0.8cm}
     \begin{subfigure}[b]{0.5\textwidth}
         \centering
        \begin{tabular}{lrr}
        \toprule
                          & \multicolumn{1}{l}{\small{Joint (\%)}} & \multicolumn{1}{l}{\small{Slot (\%)}}  \\
                          \midrule
1. DSTQA*-LastTurn            & 22.88                                & 93.53                                \\
2. SimpleTOD*-LastTurn        & 48.16                                & 96.31                                \\
3. L0P0K0-NoGraph-LastTurn   & 48.07                              & 96.88                              \\
4. L1P1K2-DSGraph-LastTurn & 49.00                              & 96.98                              \\
5. L1P1K2-DSVGraph-LastTurn & 49.25                              & 96.97                              \\
6. L1P1K3-DSVGraph-LastTurn & 49.93                              & 97.05                              \\
7. L4P4K2-DSGraph-LastTurn & \textbf{50.43}                              & \textbf{97.14}                              \\
8. L4P4K2-DSVGraph-LastTurn & 50.26                              & 97.04                              \\
9. L4P4K3-DSVGraph-LastTurn & 50.05                              & 97.04                          \\
\bottomrule
\end{tabular}
         \caption{Training with only last-turn samples.}
         \label{tab:last_turn_results}
     \end{subfigure}
     \hfill
    \caption{\textit{MultiWOZ 2.0} Dialogue State Tracking performance comparison, and ablation study. The metrics are joint accuracy (Joint) and slot accuracy (Slot) in \%. 
    GAT models are named  ``\texttt{L\{\_\}P\{\_\}K\{\_\}-[Graph\_Type]}'', for number of layers $L$, number of heads per layer $P$, and number of hops $K$ (Sec.~\ref{sec:gat_layers}).}
    \vspace{-0.5cm}
\end{table*}

We take the performance \lin{of} several recently published systems as points for comparison: 
\textbf{TRADE}~\cite{wu-etal-2019-transferable}, \textbf{DST-Picklist}~\cite{zhang2019find}, and \textbf{SUMBT+LaRL}~\cite{lee2020sumbt+}.  
These \lin{models} employ transfer learning, \lin{classification with a mixed strategy}, and reinforcement learning, respectively.
\lin{As discussed in Sec.~\ref{sec:related_work}, we also compare our model to Graph-based DSTs: \textbf{SOM-DST+SG}~\citep{zhu-etal-2020-efficient}, \textbf{GCDST}~\citep{wu-etal-2020-gcdst}, and \textbf{SST}~\cite{0002LWZT020}.}
In addition, we consider two models as most relevant baselines, and we have attempted to reproduce their results for inclusion here\footnote{The results shown in this paper might be different from what they reported. See Appendix A.1.}:
\textbf{DSTQA*}~\cite{zhou2019multi}\footnote{\href{https://github.com/alexa/dstqa}{https://github.com/alexa/dstqa}}: A bi-LSTM-based DST model utilizing a graph attention network to capture inter-slot relationships,  which motivates the architecture introduced in this paper. 

\textbf{SimpleTOD*}~\cite{NEURIPS2020_e9462095}\footnote{\href{https://github.com/salesforce/simpletod}{https://github.com/salesforce/simpletod}}: A GPT-2-based dialogue state tracker, which is similar to our base model, without graph enhancement.

\subsection{Training Regimes}
We investigate two training scenarios.   
The first approach is fully supervised at the level of individual turns, \lin{following the} common practice (e.g.~\citeauthor{NEURIPS2020_e9462095}\citeyearpar{NEURIPS2020_e9462095}).  The second approach is {\bf Sparsely-Supervised Training}, in which training is at the entire dialogue level, i.e. including only the dialogue state labels at the final turn without their intermediate states during the session, but with the previous dialogue turns included as history \revise{(shown in Fig. \ref{fig:DST_example})}.
The two components, GPT-2 and $\mathbf{GAT}$, are jointly trained.
More details are in Appendix A.2.
Under sparse supervision, the training set \lin{is} reduced from $54 ,971$ turns to only last-turn samples $7,884$ ($14.3\%$); validation utilizes only last-turn samples, as well.
Note that evaluation is performed with the standard, MultiWOZ test set ($7,372$ samples) for models trained under either regime.
For comparison, we produced the results of \textbf{DSTQA*} and \textbf{SimpleTOD*} using the same last-turn samples.
These are denoted with a ``\texttt{-LastTurn}'' suffix as in Table~\ref{tab:last_turn_results}.

\lin{We denote the configurations of $\mathbf{GAT}$ with ``\texttt{L\{\_\}P\{\_\}K\{\_\}-[Graph\_Type]}'' format, filling in number of layers $L$, number of heads per layer $P$, and number of hops $K$.}

\section{DST Performance}
\vspace{-0.1cm}

We first compare our model with baseline systems. As shown in Table \ref{tab:compare_baselines}, \texttt{L0P0K0-NoGraph}, which has no graph enhancement, achieves higher joint accuracy than most of the baseline models \lin{including the graph-based models such as GCDST and SOM-DST+SG,} setting a strong baseline for further improvement to  GPT-2-based generation.
\texttt{L4P4K2-*} models, with multiple GAT layers to encourage inter-slot information exchange, show significantly better performance.
\texttt{L4P4K2-DSGraph} achieves $54.86\%$ in joint accuracy, highest amongst these systems.

In the sparsely-supervised scenario, the performance of the baseline GPT-2 model drops to  48.07\% joint accuracy (\texttt{L0P0K0-NoGraph}, Table~ \ref{tab:last_turn_results}).
Incorporating GAT in the system (\texttt{L4P4K2-DSGraph-LastTurn}) achieves $50.43\%$ in joint accuracy, \lin{leading to} a $3\%$ degradation relative to \texttt{L0P0K0-NoGraph} fine-tuned with the full set of annotated dialogue turns.
By contrast, \texttt{DSTQA*-LastTurn}, which utilizes bi-directional LSTM modules, exhibits a sharp performance decrease to $22.88\%$ joint accuracy; we hypothesize this that the LSTM-based model can not annotate short dialogue samples well having been fine-tuned only with the last-turn samples which have relatively \lin{longer dialogue history and annotations}. 

\begin{figure*}[!h]
         \centering \includegraphics[width=0.8\textwidth]{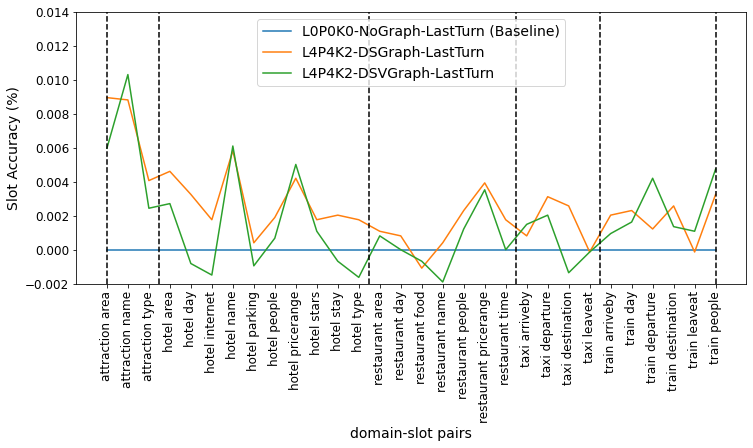}
         \caption{Slot accuracy relative to baseline (\texttt{L0P0K0-NoGraph-LastTurn}), in the serialization order for GPT-2 generation.  Domain-slot accuracy is improved, particularly for items earlier in the serialisation.}
         \label{fig:serialization_order}
         \vspace{-0.3cm}
\end{figure*}


The sparsely supervised scenario further shows the value of augmenting GPT-2 with representations derived from GATs (Table~\ref{tab:last_turn_results}).
\lin{Relative to the base system (Model 3, Table~\ref{tab:last_turn_results}), \texttt{L1P1K2-DSVGraph-LastTurn} (Model 5) improves accuracy by incorporating GAT representations in which slot nodes depend on only value nodes.}
When the number of hops is increased, slot nodes influence each other via intermediate value nodes, yielding further improvement (Models \lin{6,8}).

However, multiple GAT layers ($L=4, P=4$, Models 7,8,9, Table~\ref{tab:last_turn_results}) do not differ much in performance, showing that dependencies between slots \lin{nodes} and \lin{values} can be captured with sufficient layers \lin{(thus effectively more hops of information exchange)} and attention heads.  In particular,  although the number of hops ($K$) is relatively small,  feature passing between distant nodes can occur from layer to layer.


\revise{We summarize our findings as below:}

\revise{(1)
Through modelling values nodes, the \texttt{DSVGraph} is able to capture dependencies between slots that share values, resulting in a slight improvement over the \texttt{DSGraph} when the number of layers/hops are limited (Table \ref{tab:last_turn_results} Model 5, 6 v.s. Model 4).}

\revise{(2) With sufficient layers of GATs, \texttt{DSGraph} compensates for the lack of explicit value nodes and matches and sometimes outperforms the performance of \texttt{DSVGraph}, but this is at the cost of additional modelling complexity (comparing Table \ref{tab:last_turn_results} Model 8, 9 and Model 7).
Understanding these trade-offs will be helpful in applying these models in larger, more complex domains.}

In the following sections (Sec.~\ref{sec:analysis1}, \ref{sec:analysis2}, and \ref{sec:analysis3}), we investigate how graph modules improve the performance of the base fine-tuned GPT-2 model.

\subsection{GATs capture inter-slot dependencies}
\label{sec:analysis1}


The accuracy of each domain-slot of several models is shown in Fig.~\ref{fig:serialization_order}. The horizontal axis follows the serialization order of domain-slot pairs in the model output. 
As discussed in Sec.~\ref{sec:introduction}, when predicting \texttt{<restaurant-area>} (position 14), the causal GPT-2 model is able to condition on what has been predicted for \texttt{<attraction-area>} (position 1), but not the other direction, possibly incurring decreased performance for earlier slots.
After introducing graph modules this effect of causality is mitigated.
For example, as shown in Fig.~\ref{fig:serialization_order}, the slot accuracy of ``attraction'' domain is always boosted by graph-enhanced models (green and yellow).
We further note that these graph-enhanced models perform generally better in those intuitively correlated slots (e.g. \texttt{<hotel-pricerange>} and \texttt{<restaurant-pricerange>}).
We conclude that graph-based inter-slot dependencies are beneficial to such GPT-2-based generation models.





\subsection{GATs improve the predictions at intermediate dialogue turns}
\label{sec:analysis2}

\begin{figure}[!h]
        \includegraphics[width=0.93\linewidth]{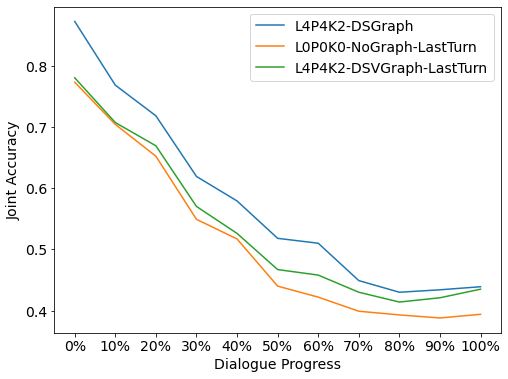}
        \vspace{-0.2cm}
        \caption{Prediction accuracy against of dialogue progress.
        Models trained with only last-turn samples can utilize GATs to retain much performance in the latter halves of dialogues ($50\%$ to $100\%$).}
         \label{fig:dialogue_progress_joint}

\end{figure}

It is important to analyze what impact the last-turn training brings to the predictions at intermediate turns, and how graph modules improve them.
A dialogue session might run for 3-4 turns to complete a single task, or up to 18 turns to complete a complex task (e.g. booking a train, taxi, and hotel in the same session).
Starting from $0\%$ (the first turn) to $100\%$ (the last turn), we report the prediction accuracy of all slots as the dialogue progresses.
As shown in Fig.~\ref{fig:dialogue_progress_joint}, the baseline model trained with all training samples (\texttt{L4P4K2-DSGraph}) shows a downward trend in prediction accuracy as the dialogue progresses.
This agrees with our observation that as dialogue progresses, the 
domain-slot prediction task becomes larger and more complex (e.g. time-related slots such as \texttt{taxi-arriveBy} \lin{are known to be difficult and} tend to appear late in a session).

Comparing \texttt{L4P4K2-DSGraph} (blue) and \texttt{L0P0K0-NoGraph-LastTurn} (\lin{yellow}), the performance throughout the dialogue sessions lags by around $5\%$ in joint accuracy.
When graph modules are introduced in models such as \texttt{L4P4K2-DSGraph-LastTurn} and \texttt{L4P4K2\\-DSVGraph-LastTurn}, the system performance in the latter half of the dialogue degrades much less.
For instance, when towards the end of dialogues (progress higher than $80\%$), the difference in joint accuracy of \lin{\texttt{L4P4K2-DSGraph} (blue) and  \texttt{L4P4K2-DSGraph-LastTurn} (brown)} is less than $2\%$.
Graph-enhanced models significantly improve the performance in the latter halves of dialogues.
A possible reason is that, as the dialogue proceeds, more values are specified and 
correlated domain-slots appear together more frequently, which enables graph modules to exploit the dependencies between slots.

\vspace{-0.1cm}
\subsection{GATs improve the predictions of correlated slots}
\label{sec:analysis3}
We investigate these inter-slot dependencies and to what extent they affect our graph models.

For every pair of value candidates under two distinct slots (e.g. \texttt{<hotel-people>:3} and \texttt{<restaurant-people>:3} form a value pair), we measure the correlation of the two values using Jaccard similarity coefficient \cite{zhang2003properties}.
Jaccard score of two sets $C_1$ and $C_2$ is defined as:
$    J(C_1, C_2)=\frac{|C_1\cap C_2|}{|C_1 \cup C_2|}$.

\begin{figure}[!h]
    \centering
    \includegraphics[width=\linewidth]{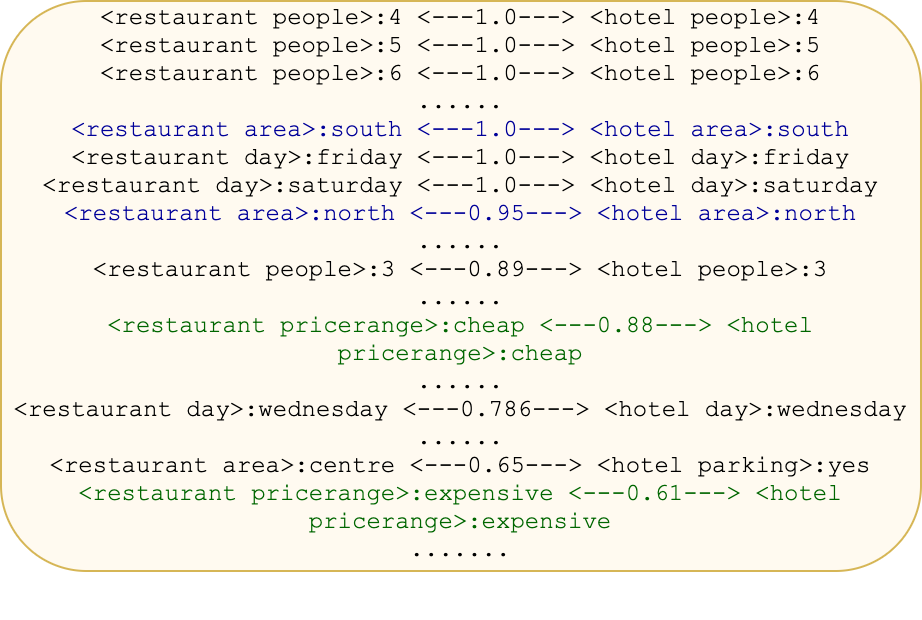}
    \vspace{-0.6cm}
    \caption{A sample of slot value pairs in the test set with their Jaccard scores. Each entry shows that values in two different slots are bridged by their Jaccord scores. Higher scores indicate stronger dependencies.}
    \vspace{-0.25cm}
    \label{fig:example_correlation}
\end{figure}

For each value pair, we flag their occurrences in the turn-level test set samples where both of their corresponding slots have non-empty annotations.
The Jaccard score is then computed from the co-occurrences of the two values. 
\lin{Further details are given in Appendix A.3.}
Intuitively, the score indicates whether the two values in the pair tend to appear together or not, which is a suitable measurement for value-level dependencies, at the same time bridging the slots to which they belong.
Note that these scores are objective values derived from the test set, without the engagement of any model.

Fig.~\ref{fig:example_correlation} shows value pairs with their Jaccard scores from the test set annotations.
There is clear evidence of dependencies in slots across domains.
For example, \texttt{<restaurant-pricerange>} and \texttt{<hotel-pricerange>} are bridged by their values (\texttt{cheap} and \texttt{expensive}) with high Jacard scores (highlighted in green).
The values of \texttt{<restaurant-area>} also aligns well with those of \texttt{<hotel-area>} (in blue).


\begin{figure}
    \centering
    \includegraphics[width=0.9\linewidth]{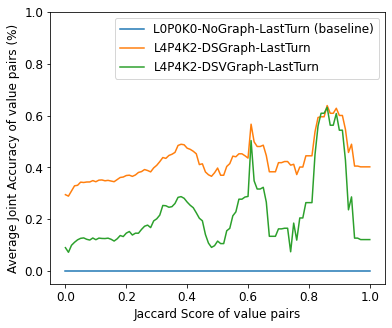}
    \vspace{-0.2cm}
    \caption{The joint accuracy (relative to baseline \texttt{L0P0K0-NoGraph-LastTurn}) changes with the Jaccard Score of value pairs. A moving window of size $0.1$ is applied to obtain the averaged joint accuracy around each Jaccard score being evaluated.}
    \vspace{-0.4cm}
    \label{fig:correlation_analysis}
\end{figure}

We run three models (as shown in the legend of Fig.~\ref{fig:correlation_analysis}) and for each value pair obtain the average pair accuracy (the success rate of correctly generating both values).
We then plot the change in  average pair accuracy (relative to baseline values) with the increasing Jaccard coefficient in Fig.~\ref{fig:correlation_analysis}. 
Compared to the baseline without GATs (blue), the graph-enhanced models (yellow and green) perform better when predicting values that have high Jaccard scores.
Specifically, \texttt{L4P4K2-DSVGraph-LastTurn} has a $0.15\%$ boost when Jaccard is around $0.2$, and it further improves the performance to $0.6\%$ at the Jaccard value of $0.88$.
Therefore, we can conclude that the graph modules enable the models to exploit the inter-slot dependencies and learn better in those highly correlated values.

\vspace{-0.1cm}
\section{Conclusion}
\label{sec:conclusion}
\vspace{-0.1cm}
We presented a novel hybrid architecture that augments GPT-2 with representations derived from Graph Attention Networks in such a way to allow causal, sequential prediction of slot values.
Our analysis shows that these graph-enhanced models mitigate some of the issues that arise in prediction with left-to-right generative models.
We also demonstrate that our model can exploit dependencies among domain-slot values, improving accuracy for systems trained with \revise{weak} supervision.



\section{Acknowledgements}
\revise{
We thank Zhilin Wang (University of Washington) for initial discussions and Qingbiao Li (University of Cambridge) for an initial implementation of graph convolution operations.
The code of this project is released on Github.\footnote{\href{https://github.com/LinWeizheDragon/Knowledge-Aware-Graph-Enhanced-GPT-2-for-Dialogue-State-Tracking}{https://github.com/LinWeizheDragon/Knowledge-Aware-Graph-Enhanced-GPT-2-for-Dialogue-State-Tracking}}}
\bibliography{anthology,custom, reference}
\bibliographystyle{acl_natbib}

\appendix
\section{Appendices}
\label{sec:appendix}


\begin{table*}[!ht]
\centering
\begin{tabular}{lccccc} 
\hline
 Sample index                     & 0      & 1         & 2         & 3         & 4        \\ 
\hline
\small{Labels for \texttt{<restaurant-pricerange>}} & none   & expensive & moderate  & expensive & moderate \\ 
\hline
 \small{Labels for \texttt{<hotel-pricerange>}}      & none   & moderate  & expensive & expensive & cheap   \\
\hline
 \small{$C_1$:\texttt{<restaurant-pricerange>:expensive}}  & ignore & 1         & 0         & 1         & 0       \\ 
\hline
 \small{$C_2$:\texttt{<hotel-pricerange>:expensive}}       & ignore & 0         & 1         & 1         & 0         \\ 
\hline
\small{dependent? $C_1\cap C_2$} & ignore & False & False & True & True \\
\hline
\end{tabular}
\caption{Example of calculating Jaccard Scores for the value pair \texttt{<restaurant-pricerange>:expensive} and \texttt{<hotel-pricerange>:expensive}. Here shows 5 possible turn-level samples to demonstrate how we flag the occurrences for the value pair.}
\label{tab:jaccard_computation}
\end{table*}

\subsection{Reproduction Details}
\label{sec:appendix:reproduction}
Dialogue state tracking performance reported in this paper are replications of published results for \textbf{DSTQA*} and \textbf{SimpleTOD*} using source code accompanying the papers describing these systems.  The asterisk indicates results found by our replication.

\begin{table}[!h]
    \centering
    \begin{tabular}{lcc}
    \toprule
         &  Joint(\%) & Slot(\%)\\
         \midrule
        \citeauthor{zhou2019multi}~\citeyearpar{zhou2019multi} & 51.44 & 97.24 \\
        DSTQA* & 52.24 & 97.28 \\
        DSTQA*-LastTurn & 22.88 & 93.53 \\
        \bottomrule
    \end{tabular}
    \caption{DSTQA Performance on MultiWOZ 2.0.}
    \label{tab:dstqa_reproduction}
\end{table}

\textbf{DSTQA*}: We used the software released by \citet{zhou2019multi}\footnote{\href{https://github.com/alexa/dstqa}{https://github.com/alexa/dstqa}} to retrain and evaluate the system with hyperparameters set as in the original code. Training ran for 300 epochs and 2 days. The best model was found at epoch 174 based on the validation accuracy of all slots.

\textbf{DSTQA*-LastTurn}: We used the same software environment as for \textbf{DSTQA*},  modified such that only the final turn of training/validation samples was used in training.
The training was run for 300 epochs and 20 hours, and the best model was found at epoch 109, after which the model exhibited overfitting and reduced performance.

\begin{table}[!h]
    \centering
    \begin{tabular}{lcc}
    \toprule
         &  Joint(\%) & Slot(\%)\\
         \midrule
        SimpleTOD* & 51.37 & 96.48 \\
        SimpleTOD*-LastTurn & 48.16 & 96.31 \\
        \bottomrule
    \end{tabular}
    \caption{Performance of SimpleTOD \cite{NEURIPS2020_e9462095} in MultiWOZ 2.0
    .}
    \label{tab:simpletod_reproduction_2.0}
\end{table}


\textbf{SimpleTOD*}: Software was downloaded from the official repository\footnote{\href{https://github.com/salesforce/simpletod}{https://github.com/salesforce/simpletod}} of SimpleTOD.   
After fixing several bugs according to the discussions in the repository, 
we evaluated this model in MultiWOZ 2.0 \cite{budzianowski2018multiwoz} for a fair comparison with our proposed models.
The best model was found by the perplexity of validation set, as recommended by the paper.

\textbf{SimpleTOD*-LastTurn}: We reduced the training data set to only final turns of dialogues as in \textbf{DSTQA*-LastTurn}, and produced the results to compare with our proposed models.





\subsection{Training Details}
\label{sec:appendix:training}
All experiments were done with a RTX3090 GPU. Fine-tuning is done with an AdamW optimizer with a linear decay learning rate for $8$ epochs ($36$ epochs for sparsely-supervised training). 
Each epoch costs around 1 hour to complete on the GPU used.
The GPT-2 component loads the pre-trained parameters of the standard model (12-layer, 768-hidden, 12-heads, 117M parameters,
OpenAI GPT-2 English model) provided by huggingface\footnote{\href{https://huggingface.co/}{https://huggingface.co/}}.
\lin{Though the GPT-2 and $\mathbf{GAT}$ are jointly trained,
the initial learning rates are $6.25\times10^{-5}$ and $8\times10^{-5}$ for two major components respectively.}
Training details can be found in our official Github repository.\footnote{\href{https://github.com/LinWeizheDragon/Knowledge-Aware-Graph-Enhanced-GPT-2-for-Dialogue-State-Tracking}{https://github.com/LinWeizheDragon/Knowledge-Aware-Graph-Enhanced-GPT-2-for-Dialogue-State-Tracking}}

\subsection{Calculation of Jaccard Scores}
\label{sec:appendix:jaccard}

Table \ref{tab:jaccard_computation} shows an example of labeling sequences of $C_1$ and $C_2$ from which Jaccard scores are computed by $J(C_1, C_2)=\frac{|C_1\cap C_2|}{|C_1 \cup C_2|}$.
With the five samples shown, the Jaccard score is $\frac{2}{4}=0.5$. The value is not high as intuitively the occurrence and absence of \texttt{<restaurant-pricerange>:expen-\\sive} does not pair well with those of \texttt{<hotel-pricerange>:expensive}.
As the number of samples increases, this score effectively reflects how a slot value depends on the other, leading to a good measurement of coreference and dependencies.




\end{document}